# Aligning Noisy Parallel Corpora Across Language Groups : Word Pair Feature Matching by Dynamic Time Warping

Pascale Fung
Kathleen McKeown
Computer Science Department
Columbia University, New York, NY 10027
pascale@cs.columbia.edu

## Abstract

We propose a new algorithm, **DK-vec**, for aligning pairs of Asian/Indo-European noisy parallel texts without sentence boundaries. The algorithm uses frequency, position and recency information as features for pattern matching. Dynamic Time Warping is used as the matching technique between word pairs. This algorithm produces a small bilingual lexicon which provides anchor points for alignment.

## 1 Introduction

While much work has already been done on the automatic alignment of parallel corpora (Brown *et al.* 1991; Kay & Röscheisen 1993; Gale & Church 1993; Church 1993; Chen 1993; Wu 1994), there are several problems which have not been fully addressed by many of these alignment algorithms. First, many corpora are noisy; segments from the source language can be totally missing from the target language or can be substituted with a target language segment which is not a translation. Similarly, the target language may include segments whose translation is totally missing from the source corpus. For example, in previous work(Church *et al.* 1993) on a Japanese/English corpus of the Awk manual, it was found that many picture files in the English text were totally missing from the Japanese version. In addition, programming examples were often quite different from one language to another. While such noisy corpora are basically parallel, they cannot be handled by traditional alignment programs which typically work on a sentence by sentence basis and thus cannot ignore a chunk of text in either source or target.

Second, most previous alignment programs have been developed and tuned for aligning European language pairs. The particular corpora used contain sentence boundaries and these form the anchor points for aligning text (e.g., (Brown *et al.* 1991) and (Chen 1993) use sentence boundaries to align French and English *Hansard*, while (Kay & Röscheisen 1993) aligns English and German *Scientific American* with sentence boundaries at the sentence level). While there has been some work on aligning English with Asian languages (Wu 1994), it also relies on sentence boundaries as anchor points. Although there are tools for automatic recognition of sentence boundaries, they are not readily available and more importantly, they cannot be applied to OCR input where punctuations are often lost in noise. Furthermore, Asian languages tend to put sentence boundaries at different places than their translation in European languages. Thus, there are more many-to-one or one-to-many sentence translations. In fact, experiments with both the length-based algorithms (Wu 1994) and with *char_align* (Church *et al.* 1993) show lower performance when used on Chinese and English. This is to be expected since the syntactic structure of Asian languages is quite different from Indo-European languages and translations are also likely to be less literal than across Indo-European pairs.

While Church has also noted in his work on aligning OCR input that the parallel corpora found in the real world are much less clean than the oft used Hansards corpora, his technique nonetheless relies on a fact which holds true of most European language pairs. *Char_align* uses cognates, or identical character sequences, as points of alignment. Such sequences are quite common in European languages (Simard *et al.* 1992), which often have the same root for a word or which the source word or phrase is left intact in the translation. This is obviously untrue for Asian/European language pairs.

Our goal is to automatically align noisy but nevertheless parallel corpora across language groups by inferring a small bilingual dictionary. This small dictionary can be used as anchor points instead of sentence boundary markers or identical character sequences for alignment. Our technique for inferring the dictionary does not rely on either identical character sequences or sentence boundaries, and thus, is ideal for aligning Asian/Indo-European language pairs as well as any noisy corpora.

In the following sections, we first describe our previous work in aligning Asian/Indo-European language pairs. We then describe our approach based on frequency, position and recency information, which we call **DK-vec**, providing motivation and details of algorithm and results.

## 2 Our Previous Work in Aligning Asian/Indo-European Language Pairs

The little work that has been done in aligning Asian/Indo-European language pairs (Wu 1994; Church *et al.* 1993) can only be applied to corpora which either contain clear sentence boundaries or identical cognates.

The algorithm we propose here is based on our previous attempt at aligning noisy corpora which don't meet these constraints using an alignment technique called K-vec (Fung & Church 1994). The idea behind K-vec was to segment both of the parallel texts into equal portions (say, K segments) and assign to each word in each of the two texts a vector with K dimensions. If the word occurs in the i-th segment for $1 <= i <= K$, then the i-th dimension of its K-vec is set to 1, otherwise 0. So word A might be $\langle 1, 0, 0, 1, 1, ...0\rangle$ and word B might be $\langle 0, 0, 0, 1, 1, ...1\rangle$.

The vector values are binary so they correspond to distribution but have some correlation with frequency[1]. Mutual information and t-scores were used as measurements on all pairs of K-vecs, thus giving a correlation measure for each pair of source and target language words. The most correlated pairs determined by these scores are assumed to be translations of each other.

Thus, K-vec is based on the assumption that if two words are translations of each other, they are more likely to occur in the same segments than two words which are not. As an example, in an extreme case, if word A occurs in three chapters, (e.g. it is present in Chapter 1, in Chapter 2, and in Chapter 10 of a document), its translation word B is most likely to be found in Chapter 1, in Chapter 2 and in Chapter 10 as well. Note that we use the notion of Chapter as segment for illustration purposes only; K-vec does not use chapter or any other *a priori* boundaries.

This technique worked well with the Canadian Hansard which was the baseline control experiment. According to the literature, K-vec is the only alignment algorithm which does not assume *a priori* language or corpus characteristics such as identical cognates or sentence boundaries. However, its performance was somehow poorer with both the Japanese/English Awk manual or the Hong Kong Hansard. It was suggested that with a small dictionary of less than 100 words, we could bootstrap the K-vec algorithm. In other words, we need to find more reliable anchor points for alignment of noisy data across language groups.

## 3 DK-vec

K-vec has shown the possibility of using frequency and position information of word pairs without *a priori* assumptions about the corpus. It is both convenient and efficient to describe such features of word pairs in vector form and use statistical scores to do pattern matching. However, K-vec segments two texts into equal parts and only compares the words which happen to fall in the same segments. This assumes a linearity in the two texts which is not likely to be accurate for Asian/Indo-European pairs. The occurrence of inserted and deleted paragraphs is another problem which leads to nonlinearity of parallel corpora.

---

[1] Alignment experiments using K-vec use words with low frequency, so the total number of presence of a word is roughly the frequency of the word in the text.

We wanted to keep the frequency and position information in vector form but we needed to capture the dynamic nature of occurrences as well. This led to the Dynamic K-vec algorithm or **DK-vec**.

DK-vec captures recency information in addition to frequency and position of words as features for matching word pairs. Dynamic Time Warping is then used as a matching technique.

### 3.1 Feature Space - Capturing Recency

We treat translation as a pattern matching task where words in one language are the templates which words in the other language are matched against. The word pairs which are considered to be most similar by some measurement are taken to be translations of each other and form anchor points for the alignment. Our task, therefore, is to find a similarity measurement which can find words to serve as anchor points.

Frequency and position of the words were used as features for similarity measurement in K-vec. In looking at the text shown in Table 1, we can see that words like *Governor* and 總督, which are translations of each other, may not necessarily occur within the same segments. To use the previous extreme case example, word A and its translation word B may occur in Chapters 1, 2 and 10 but they might not be in linearly corresponding byte segments. Chapters in the two parallel documents do not necessarily start or end at the same byte positions.

However, if we define arrival interval to be the difference between successive byte positions of a word in the text, then Table 1 shows that the arrival intervals are very similar between word pairs. Alternatively, in our previous example, although the exact byte positions of word A in chapters 1, 2 and 10 is not similar to those of word B, the fact that they each occur 3 times first at a distance of 1 chapter and then at a distance of 8 chapters is significant. This arrival interval can be regarded as the recency information of words and can be used as another feature in the pattern matching.

More concretely, the word positions of *Governor* form a vector $\langle 2380, 2390, 2463, 2565, ...\rangle$ of length 212. We compute the recency vector for *Governor* to be $\langle 10, 73, 102, 102, 91, 923, 998, ...\rangle$ with length 211. The position vector of 總督 is $\langle 90, 2021, 2150, 2158, 2238, ...\rangle$ of length 254. Its recency vector is $\langle 1931, 129, 8, 80, 91, 87, 74, 20, 728, ...\rangle$ with 253 as its length. The recency vectors of these two words have different lengths because their frequencies are different. We can compute the recency vector for any word in the bilingual texts in this way. The values of the vector are integers with lower bound of 1, and upper bound being the length of the text.

Figuratively, for each word, we have a signal such as is shown in Figure 1 with the horizontal axis the word position in the text and the vertical axis the values of the recency vectors. Note the striking similarity of the two signals which represent *Governor* in English and Chinese. The word *Bill* in Chinese is not a translation of *Governor* and its signal is clearly uncorrelated with that of *Governor*. The signal for *President* is also very different. These signal patterns of the words can be used as the features for matching to give us most similar pairs. Looking at the values of individual recency vectors, it is hard for us to tell which pairs are indeed similar to each other. However, the signals show that there is a characteristic shape of each vector which can be used for pattern matching. Just as it is hard for people tell if the spectral signals of two utterances represent the same word, so it is hard for humans to match the patterns of the recency vectors. But pattern matching has been successfully used in signal processing to match the spectral patterns for two words; we use a similar techinque to "recognize" translation of words.

Thus, DK-vec assumes that if two words are translation of each other, they are more likely to occur similar number of times at similar arrival intervals. Our task is thus to determine what distance metric to use to measure the arrival intervals and do pattern matching.

### 3.2 Pattern Matching - Dynamic Time Warping

Many pattern recognition techniques could be used to compare vectors of variable lengths in one language against vectors in the other language. We propose using Dynamic Time Warping (DTW), which has been used extensively for matching spectral signals in speech recognition tasks.

Consider the two vectors for *Governor*/總督, vector1 of length 212 and vector2 of length 254. If we plot vector2 against vector1, we get a trellis of length 212 and height 254. If the two texts were perfectly aligned translations of each other, then the interword arrival differences would be linearly proportional to the slope of the trellis, so that we could draw a straight line from the origin to the point (212, 254) to represent this correspondence between the two vectors. Since they are not, however, we need to measure the distortion of the one vector relative to the other. The way we do this is by incrementally computing the optimal path from the origin to point (212,254) as close as possible to the idealized diagonal. The divergence from the ideal gives a measure of the distortion.

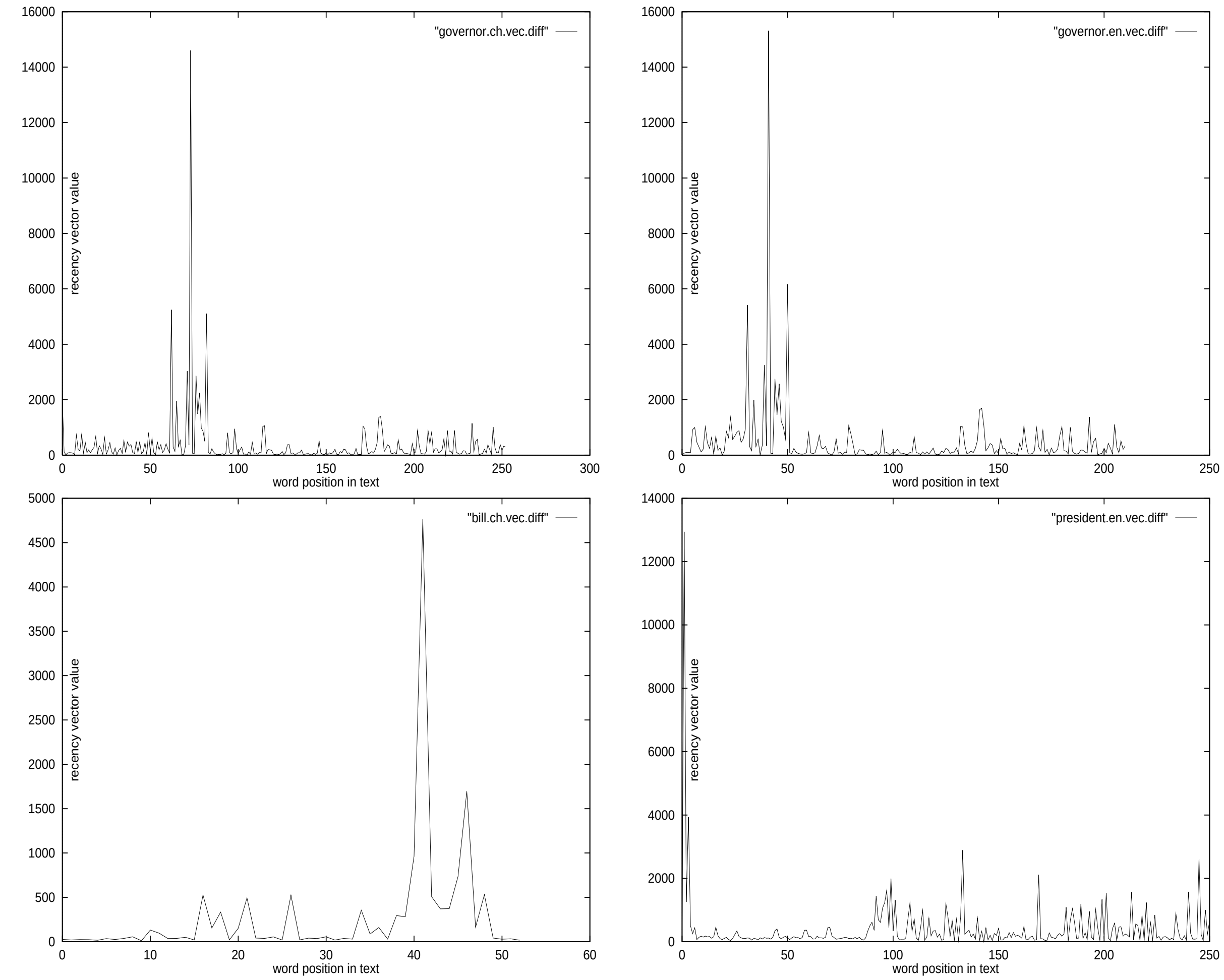

Figure 1: DK-vec signals showing similarity between *Governor* in English and Chinese, contrasting with *Bill* and *President* in English

The Dynamic Time Warping equation is

$$\begin{aligned} D(X[i:N], Y[i:N]) = \quad min\{ \quad & d(X[i], Y[i]) + D(X[i+1:N], Y[i+1:N]), \\ & d(X[i], Y[i+1]) + D(X[i+1:N], Y[i+2:N]), \\ & d(X[i+1], Y[i]) + D(X[i+2:N], Y[i+1:N])\} \end{aligned}$$

where $d(a, b) = |a - b|$

Starting from the origin, at each frame, the path is allowed to go right (1,0), up (0,1) or up-right (1,1) [2]. It is never allowed to go left because of constraints on time reversal and thus two vectors with values appearing in reverse order are never matched.

To choose among the three directions, a distortion or distance function is computed. So to decide whether the path at (i,j) should go to (i+1,j), (i,j+1) or (i+1, j+1), we compute the absolute difference between V1[i+1] and V2[j], V1[i] and V2[j+1], and between V1[i+1] and V2[j+1]. The direction is determined by the smallest absolute difference among the three pairs.

[2] This is the basic type of heuristic local continuity constraints that can be found in the literature (Rabiner & Juang 1993). Further experiments can be carried out to determine the most optimal continuity constraint for this application.

Table 1: Part of the concordances of the word *Governor* in English and Chinese

| position | arrival interval | sentence | | |
|---|---|---|---|---|
| 2380: | | MISS EMILY LAU : | Governor | - - I think I would |
| 2390: | 10 | would address you as | Governor | rather than Mr President |
| 2463: | 73 | life will be protected . | Governor | , after reading these |
| 2565: | 102 | protection of human rights , | Governor | , I could not find |
| 2667: | 102 | about other laws , | Governor | ? So will you please |
| 2758: | 91 | better to address me as | Governor | rather than Mr President |
| 3681: | 923 | MR PETER WONG : Mr | Governor | , you have adopted a |
| 4679: | 998 | I do not think , as | Governor | of Hong Kong trying |
| 5144: | 465 | IK CHI - YUEN ( in Cantonese ) : Mr | Governor | , Meeting Point welcome |
| 5439: | 295 | proposals because the | Governor | of Hong Kong and |
| 90: | | 席者 : | 總督 | 彭 定 康 先 |
| 2021: | 1931 | 定 和 以往 歷任 | 總督 | 與 本 局 所 |
| 2150: | 129 | 問 ( 譯 文 ) : | 總督 | 先生 ( 我 想 |
| 2158: | 8 | ( 我 想 稱呼 你 為 | 總督 | 先生 較 |
| 2238: | 80 | 的 社會 。 | 總督 | 先生 , 讀 畢 |
| 2329: | 91 | 滿 。 同 時 , | 總督 | 先生 , 在 保 |
| 2416: | 87 | 歡迎 。 但 | 總督 | 先生 , 其他 |
| 2490: | 74 | 未來 ? | 總督 | 答 ( 譯 文 ) : |
| 2510: | 20 | 為 你 稱呼 我 為 | 總督 | 先生 較 主 |
| 3238: | 728 | 這 是 前任 | 總督 | 領導 下 的 |

As the path is constructed, the absolute differences at each step are accumulated into a running score which is finalized when the path reaches the point (212, 254). This is the score for the pair of vectors we compared. For every word vector in language A, the word vector in language B which gives it the highest DTW score ($D(X[i:N], Y[i:N])$) is taken to be its translation. A threshold is applied to the scores to get the most reliable word pairs.

Other constraints we impose are:

- A starting point constraint - the value of the first dimension in each vector is its starting word position (minus zero). So vectors whose first value is at least half the text length apart are not considered in DTW since it means they start to occur at least half a text apart and cannot be translations of each other. This is a position constraint.

- Length constraint - the length of a vector is actually the frequency of that word minus 1. So vectors whose lengths vary by at least half the length of the text represent cases where one word occurs at least twice as often as the other word; thus they cannot be translations of each other. They are not considered for DTW either. This is a frequency constraint.

DTW assumes the translation of a word in language A is just its “distortion” in language B. DTW “warps” such “distorted” words back to their “original” form. So *Governor* is just a distortion of 總督.

### 3.3 Algorithm

The algorithm is as follows:

1. For each word in the word list of each text, find its word positions in the text. For *Governor*, the positions are $\langle 2380, 2390, 2463, 2565, 2667, ... \rangle$

2. For each word, compute the difference between arrival times and assign to that word this new vector. Recency vector for *Governor* is $\langle 10, 73, 102, 102, ... \rangle$. All words have a vector of *variable* length (depending on its freqency and position in the text). The values of the vector are integers with minimum of 1, and maximum the length of the text.

3. Filter the pairs of recency vectors, removing pairs which do not meet the stated constraints. If the first time the two words occur is over half a text apart, this pair is removed. If one vector has length smaller than half of the length of the other vector, it means the frequency of this one word is less than half of the frequency of the other word. This pair is also filtered out as an improbable translation.

4. For all pairs of words whose recency vectors survive the filtering process, compute the distance function between the words using Dynamic Time Warping. The distance function in DTW is taken to be the absolute difference value of the corresponding dimensions of the two vectors.

5. In order to determine the *closest* pair of words, sort the word pairs according to their DTW score. A small DTW score represents close correlation of two words. For pairs of words whose DTW scores are below certain threshold, we get a small bilingual dictionary which can be used as anchor points in alignment.

## 4 Experimental Results

Experiments were done on part of the HKUST parallel corpus(Wu 1994) which contains transcripts of Hong Kong Legislative Council debates. This part of the corpus contains about 700K of data in total.

The algorithm was tried on words with frequency 10 to 300 in order to capture the open-class words since their frequency typically falls in this range in English (Fung & Church 1994). An open-class word is likely to have a more characteristic signal because different words of the same class are likely to be used in distinct contexts. In contrast, words such as determiners (e.g., the demonstratives "this" or "that") are likely to be used in very similar contexts (e.g., before a wide variety of adjectives and nouns) and thus, would have indistinguishable signals. In fact, our experiments show that nouns yield some of the best results. The program was run in both directions, i.e., Chinese to English and English to Chinese. The resultant lists were slightly different but both give us translated pairs of words. They can be combined to give an optimal list.

As an example, we show the first 42 pairs of Chinese and English words which have the lowest DTW scores in Table 2. The list shows 32 correctly translated pairs of words. Some of them are translations of collocated words. (e.g. 一氧化碳 is translated to both *carbon* and *monoxide* because it is one word in Chinese but two in English). The other 10 pairs of words (shown with a '$*$') are close neighbors in the text. These initial bilingual word pairs can act as anchor points for a rough alignment.

The left graph of Figure 2 is a scaled graph showing the DTW pattern matching result. Every dot at *i,j* represents the fact that input segment at position *i* from text 1 is determined to be most likely correlated with input segment at position *j* in text 2. We can visualize a line across the diagonal. We used a standard dynamic programming algorithm to trace these dots, and the right graph of Figure 2 shows the final alignment path. If text 1 was identical to text 2 or if they were perfectly aligned to each other, the line would be a perfect diagonal. Our alignment path follows a rough diagonal line diverging from the ideal.

Table 2: Part of the DK-vec bilingual lexicon output

| DTW | score | English word | Chinese word | DTW | score | English word | Chinese word |
|---|---|---|---|---|---|---|---|
| 2 | | Ltd | 有限公司 | 28 | | air | 空氣 |
| 6 | | monoxide | 一氧化碳 | 28 | | Deputy | 副 |
| 6 | | carbon | 一氧化碳 | 29 | * | clause | 監管 |
| 6 | | P | J | 29 | * | Tunnel | 業主 |
| 6 | | J | J | 31 | | waters | 船隻 |
| 8 | | E | E | 31 | * | posts | 發牌 |
| 9 | | caisson | 沉箱 | 37 | | Basic | 基本法 |
| 9 | | Travel | 旅遊 | 37 | | ? | ? |
| 10 | * | licence | 旅遊 | 38 | | centres | 運輸 |
| 11 | | HONOURABLE | J | 39 | | McGREGOR | 覺 |
| 14 | * | outbound | 租金 | 41 | | Law | 基本法 |
| 15 | | tunnels | 沉箱 | 43 | | Bill | 草案 |
| 16 | | Industry | 旅遊 | 45 | | $ | 元 |
| 18 | | storage | 儲存 | 49 | * | imported | 管制 |
| 19 | | gas | 儲存 | 49 | | Governor | 總督 |
| 19 | * | amusement | 監管 | 52 | * | Building | 試驗 |
| 19 | | TRANSPORT | 牌照 | 53 | | Second | 二讀 |
| 22 | * | buildings | 儲存 | 57 | | Clause | 出租 |
| 24 | | C | C | 58 | | President | 副 |
| 25 | | O | E | 73 | * | quality | C |
| 28 | | vessels | 船隻 | 76 | | SECURITY | 保安 |

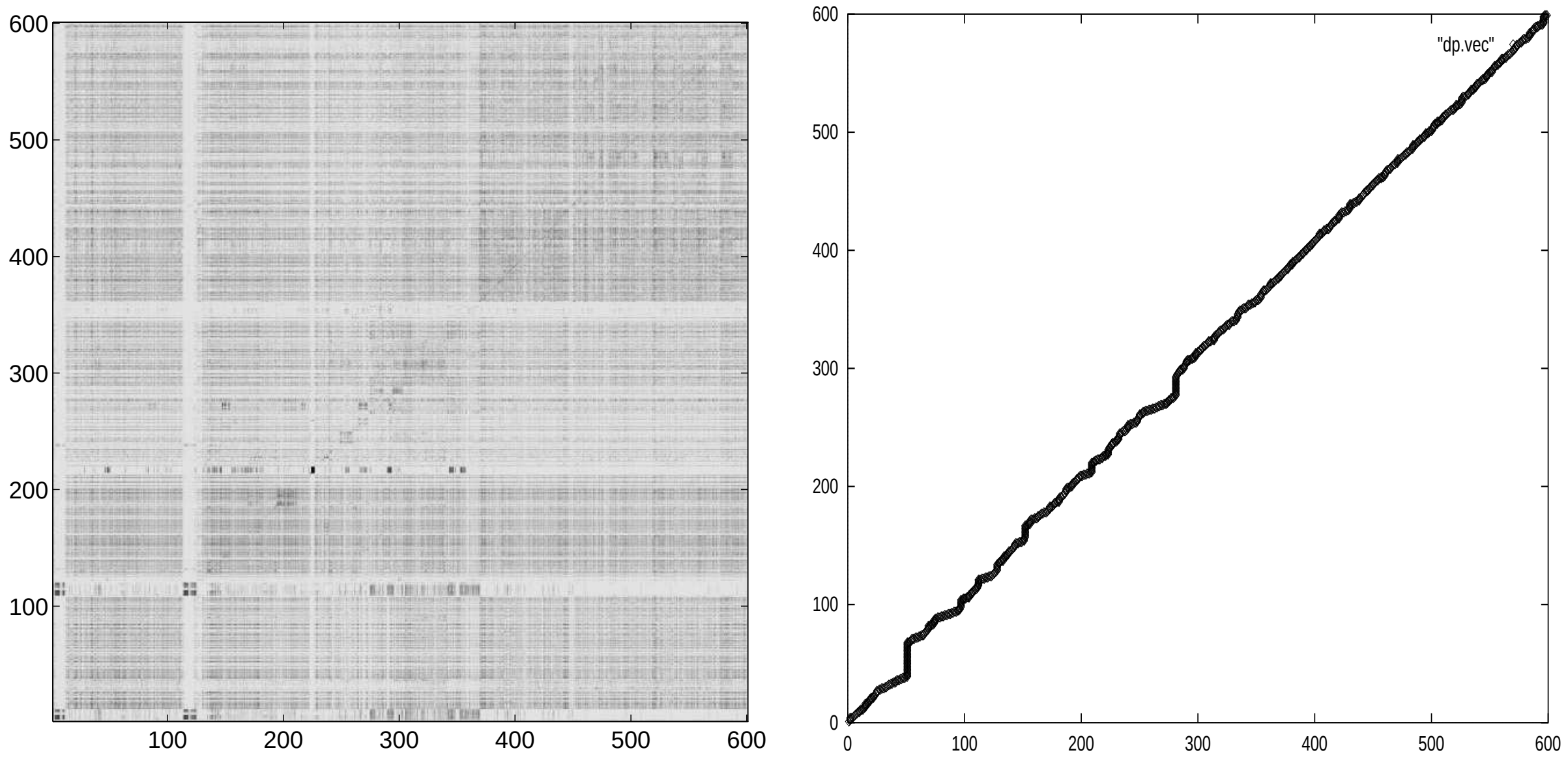

Figure 2: DTW output and the final alignment path tracing the anchor points

# 5 Conclusion

DK-vec captures the frequency, position and recency information of words in parallel texts to be used in alignment. It produces a small dictionary to be used as anchor points for alignment. Since DK-vec does not assume sentence boundaries, it can be applied to OCR input. It works on corpora with very different character sets and across quite different language groups. Most importantly, it can be applied to align noisy corpora with inserted and deleted segments of texts. Because of these properties, the choice of source and target languages to be aligned by DK-vec can be arbitrary. DK-vec can be applied to a vast range of noisy parallel corpora in any languages.

DK-vec differs from other pattern matching methods used in machine translation in that its features are explicit and tangible. The frequency, position and recency parameters are obvious and easy to manipulate. This avoids the local maxima problem in hidden parameter tuning in EM-based algorithms. In the absence of sentence boundaries, DK-vec can provide a reliable initializing point for further iterations in EM such as the algorithm used in (Wu & Xia 1994) for obtaining a full bilingual lexicon. We hope to incorporate more features such as second differences of recency information, statistical features of the vectors, or additional linguistic information into the pattern matching processing and extend DK-vec to be a full word translation algorithm.

We also noticed that most of word pairs in the bilingual lexicon produce by DK-vec are noun phrases including some word members of a collocation. This agrees with our intuition that whereas most function words, verbs, adjectives, etc, can have multiple translations in another language, proper nouns, technical terms and some nouns have fairly consistent translations in any given text. Their frequency, position and recency patterns are therefore more similar to those of their translations. It is possible that DK-vec be extended to produce noun phrase translations or even collocation translations.

## 6 Acknowledgment

We would like to thank Kenneth Church, William Gale, Chin-Hui Lee and Dekai Wu for useful discussions and pointers, and AT&T Bell Labs and the Hong Kong University of Science & Technology for use of software tools and facilities. We would also like to thank the anonymous reviewers for their very useful comments. This work was supported jointly by ARPA and ONR under contract N00014-89-J-1782 and by NSF GER-90-24069.